\documentclass[journal]{IEEEtranTIE}

\usepackage{graphics} 
\usepackage{epsfig} 
\usepackage{times} 
\usepackage{amsmath} 
\usepackage{amssymb}  
\usepackage{physics}
\usepackage{cite}
\usepackage{cleveref}
\usepackage{tabularx}
\usepackage{multirow}

\usepackage{algorithm}
\usepackage{algorithmic}

\usepackage{booktabs}
\usepackage{xcolor}
\usepackage{makecell}
\usepackage{lettrine}
\usepackage{fancyhdr}
\usepackage{graphicx}
\usepackage{subcaption}
\usepackage{url}
\usepackage[font=footnotesize]{caption}
\begin{document}

\title{PRTL: A Proxy Return Tendon Loop for Configuration-Dependent Hysteresis Compensation in Tendon-Sheath Mechanisms}

\author{
	\vskip 1em
	Myeongbo Park$^{\ast}$, Junhyun Park$^{\ast}$, and Minho Hwang$^{\dagger}$
	\thanks{$^{\ast}$Equal contribution as first authors. $^{\dagger}$Corresponding author.}
	\thanks{This work was supported by the Industrial Strategic Technology Development Program (ISTDP), funded by the Ministry of Trade, Industry and Energy (RS-2024-00443054); the Basic Science Research Program through the National Research Foundation of Korea (NRF), funded by the Ministry of Education (RS-2025-25420118); the Korea Medical Device Development Foundation grant funded by the Korean government (2420024769, RS-2026-25540191); the Korea Health Technology R\&D Project through the Korea Health Industry Development Institute (KHIDI), funded by the Ministry of Health \& Welfare (RS-2026-25503221); and the InnoCORE program of the Ministry of Science and ICT (26-InnoCORE-01 and 26-InnoCORE-08).}
	\thanks{M. Park and M. Hwang are with the Department of Robotics and Mechatronics Engineering, DGIST, Daegu, Republic of Korea (e-mail: \{qkraudqh23, minho\}@dgist.ac.kr).}
	\thanks{J. Park conducted this work with the Department of Robotics and Mechatronics Engineering, DGIST, Daegu, Republic of Korea, and is currently affiliated with the AI Research Lab, DEEPNOID, Seoul, Republic of Korea (e-mail: sean05071@dgist.ac.kr).}
}

\maketitle

\begin{abstract}
Tendon–sheath mechanisms (TSMs) enable remote tension transmission along curved paths, but their hysteresis varies with the routing configuration, limiting the effectiveness of fixed-parameter compensation. Updating the transmission model often requires distal tension measurements or additional shape or motion sensing, which can complicate system integration. This paper proposes the Proxy Return Tendon Loop (PRTL), an auxiliary tendon loop routed alongside the actuation TSM and returned through a distal pulley, allowing both loop-end tensions to be measured proximally. A 10-s probing motion characterizes the PRTL input–output tension relationship at the current configuration without driving the actuation tendon. A closed-form analytical mapping accounts for the double-pass sheath routing and calibrated pulley transmission to estimate the pulling and releasing propagation gains and biases of the actuation TSM. These parameters are used for feedforward hysteresis compensation in both propagation and backlash phases. Across eight bending configurations, the mapped models yield mean output-tension prediction root-mean-square errors (RMSEs) of 0.55 N and 0.49 N for pulling and releasing, respectively, relative to directly identified actuation TSM models. In tension-tracking experiments across two planar and two spatial configurations, the proposed method achieves mean RMSEs of 0.59 N and 0.79 N under triangular and random references, respectively, reducing errors by 85.7\% and 84.8\% compared with uncompensated control. These results demonstrate that PRTL-based parameter estimation supports tension-tracking compensation in a single actuation TSM without distal tension sensing for parameter estimation.

\end{abstract}


\section{Introduction}
\label{sec:intro}

Flexible endoscopic surgical robots provide access to internal organs through natural orifices, enabling minimally invasive procedures without external incisions~\cite{Continuum,K-FLEX,MASTER,ESD,PETH_HWANG}. These robots use tendon--sheath mechanisms (TSMs) to transmit actuation from proximal motors to distal joints. TSMs enable compact distal instruments with low inertia and allow flexible routing along curved insertion paths~\cite{KatoRobot,FIORA}.

Despite these advantages, TSMs exhibit configuration-dependent hysteresis that reduces control precision. As the insertion tube conforms to the anatomical path, changes in tendon routing alter the local curvature and accumulated bending angle of the TSM~\cite{T1_2,TIE_Sensorless}. These geometric changes affect tendon--sheath friction and tension-dependent tendon elongation, thereby changing the hysteretic relationship between the proximal input and distal response~\cite{kimhansol}. Accurate control therefore requires identifying the transmission characteristics at the current TSM configuration and incorporating them into the control strategy.

Existing configuration-dependent hysteresis compensation approaches use fiber Bragg grating (FBG) shape sensing, distal tension measurements, or visual observation of distal joint motion~\cite{TSMCOMP_FBG,TIE_Sensorless,HONG_RAL}. FBG sensors are fragile and difficult to integrate. Distal force sensors must fit within the limited space already occupied by tendons and joint mechanisms, making their integration challenging while maintaining a compact instrument design. Visual probing relies on visibility of the distal instrument or marker, which can be interrupted by occlusion, and requires predefined instrument motions that may be difficult to execute in confined anatomical spaces. These limitations motivate identifying configuration-dependent transmission parameters from proximal measurements without moving the actuated joints. Related approaches are reviewed in Section~\ref{sec:related_work}.

This study proposes the Proxy Return Tendon Loop (PRTL), an auxiliary tendon loop routed alongside the actuation TSM and returned through a distal pulley (Fig.~\ref{fig:sstl_design}). Both loop-end tensions are measured proximally during a 10-s probing motion without driving the actuation tendon. A closed-form mapping converts the identified PRTL parameters into actuation TSM transmission parameters for feedforward hysteresis compensation.

The contributions of this work are:
\begin{enumerate}
\item A double-pass PRTL structure that captures configuration-dependent tension transmission through proximal sensing, without distal force sensors, shape sensors, or motion of the actuated joints.
\item A PRTL input--output tension model that yields a closed-form mapping from the identified PRTL parameters to the actuation TSM transmission parameters.
\item A feedforward controller that uses the mapped parameters to compensate for hysteresis in both propagation and backlash phases. Validation across two planar and two spatial configurations shows mean tension-tracking root-mean-square error (RMSE) reductions of 85.7\% and 84.8\% under triangular and random references, respectively, compared with uncompensated control.
\end{enumerate}

\section{Related Work}
\label{sec:related_work}

\subsection{Hysteresis Modeling and Compensation}
Existing methods compensate for TSM hysteresis through transmission modeling or feedback from the distal response.

Model-based approaches describe TSM friction and hysteresis using the capstan equation, LuGre models, or Bouc--Wen models~\cite{T1_1,T1_2,T1_3,T3_1,T3_2,T3_3,T3_4,T3_5}. Extensions incorporating tendon elongation account for the input--output position relationship~\cite{T2_1,T2_3}. Data-driven approaches use LSTM or TCN architectures to learn the effects of tendon--sheath friction and backlash~\cite{T4_1,T4_2,T4_3,mb_iros}. Hybrid approaches combine neural networks with the Preisach model~\cite{T4_4} or use the Koopman operator to represent nonlinear tendon dynamics~\cite{T4_6}. Markerless pose estimation has also been used for visual servoing to compensate for TSM-induced motion errors~\cite{VISUAL_SERVO}. Applying these approaches across different TSM bending configurations also requires accounting for the associated changes in transmission characteristics.

\begin{figure}[t!]
    \centering
    \includegraphics[width=\columnwidth]{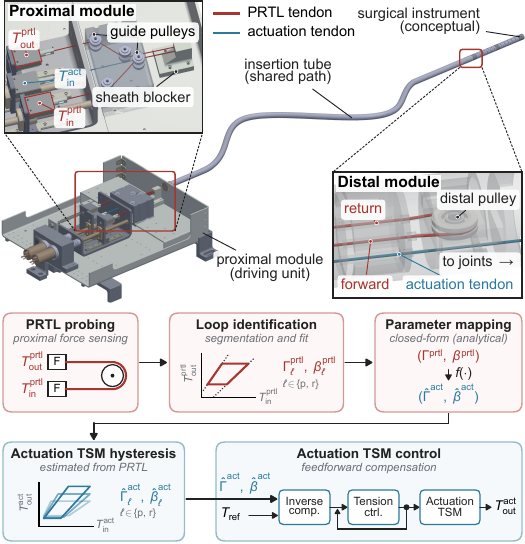}
    \caption{Overview of the PRTL system and compensation workflow. Top: shared tendon routing and enlarged views of the proximal and distal modules. Bottom: flowchart of PRTL-based configuration-dependent hysteresis compensation, from probing and parameter identification to mapping and feedforward control.}
    \label{fig:overview_concept}
    \vspace{-10pt}
\end{figure}

\subsection{Compensation for Configuration-Dependent TSM Hysteresis}
To account for changes in TSM bending configuration, prior compensation methods use shape sensing, tension measurements, or visual observation to identify the current transmission characteristics.

FBG sensors measure the TSM bending shape, which is then used to predict and compensate for the shape-dependent hysteresis~\cite{TSMCOMP_FBG}. Fiber fragility and the need for precise alignment and bonding complicate sensor integration. Lu et al.~\cite{TIE_Sensorless} proposed a double-TSM structure in which a second TSM equipped with proximal and distal tension sensors estimates the bending angle for feedback control of the main actuation TSM. Dense tendon routing and limited distal space make distal force-sensor integration difficult in endoscopic manipulators. Hong et al.~\cite{HONG_RAL} estimated the accumulated bending angle using an equivalent-circle model and identified the backlash phase by tracking a distal marker with an endoscopic camera. Such probing requires marker visibility and predefined joint motions, which can be difficult to perform in confined anatomical spaces.

Closely related studies estimate bending from proximal tendon displacement in wearable hand robots, whose short TSM paths (approximately 15~cm) exhibit little backlash~\cite{T5_2,T5_3}. For approximately 2-m endoscopic TSMs, accumulated tendon deformation and backlash make configuration estimation from displacement alone more difficult.

These limitations motivate identifying configuration-dependent TSM transmission parameters through proximal tension measurements during probing, rather than relying solely on tendon displacement differences. Routing both ends of a tendon loop to the proximal side enables tension-based identification without distal sensors, visual observation of instrument motion, or FBG shape sensing.

\begin{figure}[t!]
    \centering
    \includegraphics[width=\columnwidth]{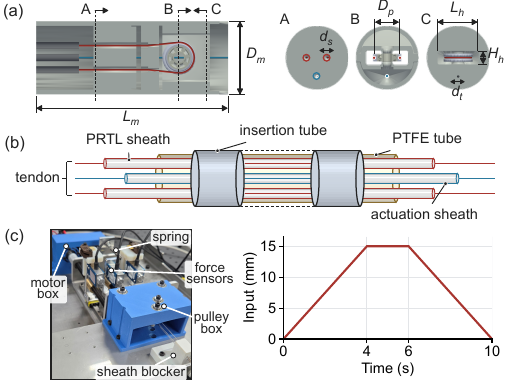}
    \caption{PRTL components and sheath routing. (a) Distal pulley module with sectional views A--C; dimensions are given in Table~\ref{tab:module_spec}. (b) PRTL and actuation sheaths within the shared PTFE tube. (c) Driving unit components and the 10-s probing trajectory.}
    \label{fig:sstl_design}
    \vspace{6pt}
    {\makeatletter\def\@captype{table}\makeatother\footnotesize
    \caption{Specifications of the Distal Pulley Module}
    \label{tab:module_spec}
\begin{tabular*}{\columnwidth}{@{\extracolsep{\fill}}lccc}
\hline
\textbf{Parameter} & \textbf{Symbol} & \textbf{Value} & \textbf{Unit} \\
\hline
Module length          & $L_m$ & 40.5 & mm \\
Module diameter        & $D_m$ & 15.5 & mm \\
Pulley diameter        & $D_p$ & 6.0 & mm \\
Tendon hole diameter   & $d_t$ & 0.7 & mm \\
Sheath blocker diameter & $d_s$ & 1.35 & mm \\
Hollow length          & $L_h$ & 10 & mm \\
Hollow height          & $H_h$ & 3.2 & mm \\
\hline
\end{tabular*}}
    \vspace{-14pt}
\end{figure}

\section{System Description and Parameter Identification}
\label{sec:system_design}
Fig.~\ref{fig:overview_concept} summarizes the overall approach: PRTL transmission parameters are identified from proximal tension measurements (Section~\ref{subsec:Parameter_Identification}) and mapped to the actuation TSM transmission parameters (Section~\ref{sec:mapping}) for feedforward compensation (Section~\ref{sec:control}).
This section describes the PRTL hardware and probing procedure, followed by tension modeling, hysteresis loop segmentation, and transmission parameter identification.

\subsection{PRTL Structure and Probing}
\label{subsec:sstl_design}
The PRTL routes a proxy tendon through the insertion tube and back to the proximal driving unit, where both its input and output tensions ($T_{\mathrm{in}}^{\mathrm{prtl}}$ and $T_{\mathrm{out}}^{\mathrm{prtl}}$) are measured.

The PRTL comprises two TSMs joined by the distal pulley: the forward TSM connected to the motor and the return TSM connected to the spring (Fig.~\ref{fig:sstl_design}). Module dimensions are listed in Table~\ref{tab:module_spec}. The two PRTL TSMs and the actuation TSM share a PTFE tube inside the insertion tube (Fig.~\ref{fig:sstl_design}(b)). This shared routing constrains the three TSMs to follow similar bending paths as the insertion tube deforms. All three use identical mechanical specifications, with 0.45-mm tendons and sheaths with outer/inner diameters of 1.1/0.6~mm. These shared geometric and mechanical characteristics provide the basis for the parameter mapping in Section~\ref{sec:mapping}.

The driving unit comprises a motor, a spring, and two force sensors (Fig.~\ref{fig:sstl_design}(c)). To characterize tension transmission during pulling and releasing at the current configuration, a 10-s trapezoidal displacement trajectory is applied to the PRTL while $T_{\mathrm{in}}^{\mathrm{prtl}}$ and $T_{\mathrm{out}}^{\mathrm{prtl}}$ are recorded. The identified parameters can be reused during instrument operation while the insertion-tube configuration remains stable. Probing is repeated to update the parameters after a configuration change.

\subsection{Tension Transmission Model}
\label{subsec:friction_model}
\begin{figure}[t]
    \centering
    \includegraphics[width=\columnwidth]{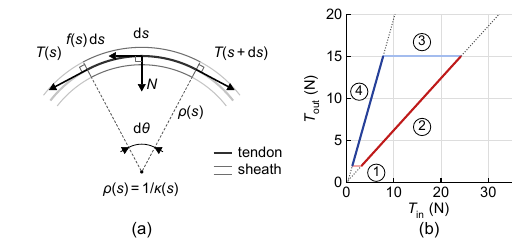}
    \caption{Tendon--sheath tension transmission. (a) Tendon element with curvature radius $\rho(s)=1/\kappa(s)$ and local bending angle $d\theta=\kappa(s)\,ds$. (b) Four-phase hysteresis loop: 1) pulling backlash (PB), 2) pulling propagation (PP), 3) releasing backlash (RB), and 4) releasing propagation (RP).}
    \vspace{-10pt}
    \label{fig:tsm_hysteresis}
\end{figure}

The TSM input--output tension relationship is modeled to define the gain and bias used for parameter identification and mapping. The hysteresis loop comprises four phases: pulling backlash (PB), pulling propagation (PP), releasing backlash (RB), and releasing propagation (RP) (Fig.~\ref{fig:tsm_hysteresis}(b)). Output tension varies with input tension during propagation and is approximated as constant during backlash following a motion reversal.

For a tendon sliding within a curved sheath (Fig.~\ref{fig:tsm_hysteresis}(a)), the ideal Coulomb friction magnitude per unit length is $f(s)=\mu T(s)\kappa(s)$, where $s$ is arc length from the input to the output, $\mu$ is the friction coefficient, and $\kappa(s)$ is the local curvature. The quasi-static tension balance gives
\begin{align}
\label{eq:local_balance}
dT(s) = -\mu \zeta\, T(s)\,\kappa(s)\, ds.
\end{align}
Here, $\zeta=+1$ for pulling propagation, when the tendon slides toward the input end, and $\zeta=-1$ for releasing propagation, when it slides toward the output end. Integrating over the sheath length $L$ gives the ideal relation $T_{\mathrm{out}}=T_{\mathrm{in}}e^{-\mu\zeta\Phi}$, where $\Phi=\int_0^L\kappa(s)\,ds$ is the accumulated bending angle. To include radial and axial corrections, we adopt the corrected capstan model of Lu et al.~\cite{TIE_Sensorless}:
\begin{equation}
\label{eq:capstan_pass}
T_{\mathrm{out}} = T_{\mathrm{in}}\,e^{-(\mu\zeta+R')\Phi} + A\bigl(e^{-(\mu\zeta+R')\Phi}-1\bigr).
\end{equation}
Here, $R$ and $A$ are radial and axial correction terms, respectively, and $R'=R/N$, with $N$ denoting the normal force. The effective coefficients $R'$ and $A$ are treated as constant within each propagation phase.

The propagation relation is represented by a gain $\Gamma$ on the input tension $T_{\mathrm{in}}$ and an additive bias $\beta$. With output tension approximated as constant during backlash, the tension relationship is written as
\begin{equation}
\label{eq:Tension_base}
T_{\mathrm{out}}(t) =
\begin{cases}
\Gamma\,T_{\mathrm{in}}(t)+\beta, & \text{(Propagation)}, \\
T_{\mathrm{out}}(t_{\mathrm{rev}}), & \text{(Backlash)},
\end{cases}
\end{equation}
where $t_{\mathrm{rev}}$ is the time of the most recent reversal in the input direction, at which backlash begins.

The propagation gain $\Gamma$ and bias $\beta$ are identified separately for pulling and releasing from measured data (Section~\ref{subsec:Parameter_Identification}) and used for parameter mapping (Section~\ref{sec:mapping}).

\subsection{Hysteresis Loop Segmentation and Parameter Identification}
\label{subsec:Parameter_Identification}
This subsection describes how measured input--output tension loops are segmented to identify the pulling and releasing propagation parameters of the PRTL and actuation TSM.

The raw tension signals ($T_{\mathrm{in}}$, $T_{\mathrm{out}}$) are smoothed using a zero-phase Butterworth filter to suppress sensor noise, yielding $\tilde{T}_{\mathrm{in}}$ and $\tilde{T}_{\mathrm{out}}$. Turning points, where $\tilde{T}_{\mathrm{in}}$ switches between increasing and decreasing, divide the loop into pulling (increasing input tension) and releasing (decreasing input tension) segments. Within each segment, the backlash phase is identified where $\tilde{T}_{\mathrm{out}}$ remains within a threshold $\delta_{\mathrm{bl}}$ of its segment-initial value. We set $\delta_{\mathrm{bl}}$ to 0.3~N for the actuation TSM and 0.05~N for the PRTL. The remaining samples are assigned to the corresponding propagation phase.

\begin{figure}[t]
    \centering
    \includegraphics[width=\columnwidth]{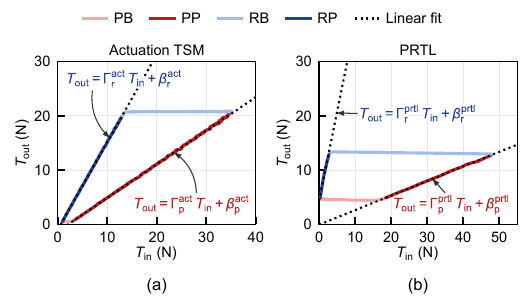}
    \caption{Loop segmentation and parameter identification for (a) the actuation TSM and (b) the PRTL. Colors distinguish the four hysteresis phases; dotted lines show the linear fits to the pulling and releasing propagation phases.}
    \vspace{-10pt}
    \label{fig:param_extraction}
\end{figure}

A separate linear model is then fitted to each pulling and releasing propagation phase using the relation in~\eqref{eq:Tension_base}. Let $\mathcal{P}_{\ell}$ denote the propagation sample indices for direction $\ell\in\{\mathrm{p},\mathrm{r}\}$, where $\mathrm{p}$ and $\mathrm{r}$ denote pulling and releasing, respectively. The gain and bias are obtained by least-squares regression:
\begin{align}
\label{eq:ls_fit}
\left(\Gamma_{\ell},\, \beta_{\ell}\right)
= \arg\min_{\Gamma,\, \beta} \sum_{n \in \mathcal{P}_{\ell}}
\left(\tilde{T}_{\mathrm{out}}[n] - \Gamma\, \tilde{T}_{\mathrm{in}}[n] - \beta\right)^{2}.
\end{align}
Applying this procedure to both systems yields $\psi^{s}=\{\Gamma_{\mathrm{p}}^{s},\,\beta_{\mathrm{p}}^{s},\,\Gamma_{\mathrm{r}}^{s},\,\beta_{\mathrm{r}}^{s}\}$, where $s\in\{\mathrm{act},\mathrm{prtl}\}$ identifies the actuation TSM or PRTL. Fig.~\ref{fig:param_extraction} shows the segmented loops and fitted propagation models.

\section{Hysteresis Parameter Mapping}
\label{sec:mapping}
This section derives an analytical mapping to estimate the propagation parameters of the actuation TSM using the proximally measured PRTL loop parameters. The propagation relation in~\eqref{eq:Tension_base} represents the input--output tension relationship using a gain and a bias. Using this representation, the relationship between $T_{\mathrm{in}}^{\mathrm{prtl}}$ and $T_{\mathrm{out}}^{\mathrm{prtl}}$ is modeled in terms of the gains and biases of the individual TSM segments and pulley stages. The following derivation applies separately to the pulling and releasing directions. The mapping and its component models are formulated under the following assumptions:

\noindent 1) {\fontfamily{lmr}\selectfont\itshape Common bending geometry across the TSMs:} The actuation TSM and the two PRTL sheath passes are co-routed within the same PTFE bundle and are assumed to follow similar paths with approximately the same accumulated bending angle $\Phi$.

\noindent 2) {\fontfamily{lmr}\selectfont\itshape Common mechanical parameters:} With identical tendons and sheaths, the three sheath passes are modeled using a common friction coefficient $\mu$ and effective correction parameters $R'=R/N$ and $A$ within the operating range.

\noindent 3) {\fontfamily{lmr}\selectfont\itshape Linear algebraic tension relation:} Within each propagation phase, each stage is approximated by an affine tension relation, and the stage relations are composed algebraically.

\begin{figure}[t!]
    \centering
    \includegraphics[width=\columnwidth]{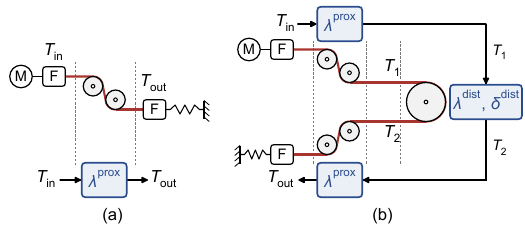}
    \caption{Schematics of the pulley identification setups with tendon only and their gain--bias cascades. (a) Proximal drive-pulley pair. (b) Combined proximal and distal pulleys: two proximal pairs and one distal return pulley. Intermediate tensions $T_1$ and $T_2$ define the cascade in \eqref{eq:pulley_cascade}.}
    \label{fig:pre_params}
    \vspace{-8pt}
\end{figure}

\begin{figure}[t!]
    \centering
    \includegraphics[width=\columnwidth]{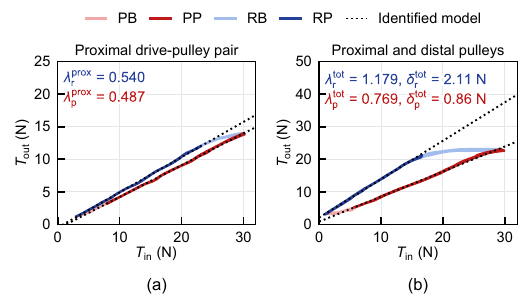}
    \caption{Measured pulley transmission and identified models. (a) Proximal drive-pulley pair. (b) Combined proximal and distal pulleys. Colors indicate hysteresis phases; dotted lines show the pulling and releasing propagation models.}
    \label{fig:pulley_id}
    \vspace{-8pt}
\end{figure}

\subsection{Pulley Parameter Identification}
\label{sec:pre-def}

Unlike the actuation TSM, the PRTL uses proximal pulleys to align the tendon path from the motor with the centerline of the insertion tube and a distal pulley to redirect the tendon along the return path, as shown in Fig.~\ref{fig:overview_concept}. To investigate nonlinear tension transmission associated with the pulleys, two bench setups were constructed with only the tendon routed through the pulleys, without sheaths, as shown in Fig.~\ref{fig:pre_params}(a) and (b). Both experiments were conducted under the same spring load as the PRTL system. The measurements revealed additional gain and bias effects in both setups (Fig.~\ref{fig:pulley_id}(a) and (b)). These observations motivated the identification of effective pulley transmission coefficients to account for these effects in the parameter mapping. In the combined setup consisting of two proximal pulley pairs and one distal return pulley (Fig.~\ref{fig:pre_params}(b)), the measured input--output relationship is represented by the total gain $\lambda^{\mathrm{tot}}$ and bias $\delta^{\mathrm{tot}}$:
\begin{equation}
\label{eq:pulley_tot}
T_{\mathrm{out}} = \lambda^{\mathrm{tot}} T_{\mathrm{in}} + \delta^{\mathrm{tot}}.
\end{equation} 

\begin{figure*}[t]
    \centering
    \includegraphics[width=\textwidth]{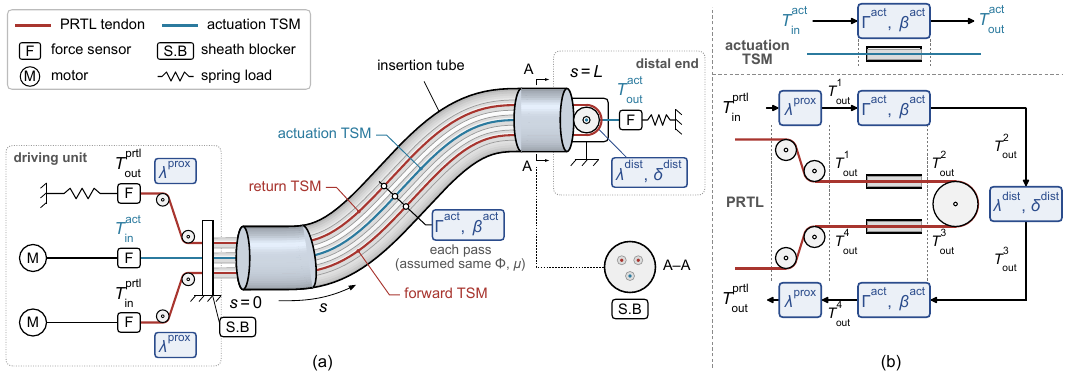}
    \caption{PRTL tension model. (a) Shared routing of the PRTL (red) and actuation TSM (blue). (b) Gain--bias cascade of the two PRTL sheath passes, two proximal pulley pairs, and distal return pulley. Both sheath passes use the actuation TSM parameters under Assumptions~1 and~2. Intermediate tensions $T_{\mathrm{out}}^{1}$--$T_{\mathrm{out}}^{4}$ are defined in \eqref{eq:stage_chain}.}
    \vspace{-10pt}
    \label{fig:prtl_model}
\end{figure*}

To determine the individual stage coefficients, the proximal gain was first estimated from the proximal-only setup (Fig.~\ref{fig:pre_params}(a)). The measured input--output relationship was approximately proportional, with a small bias relative to the operating tension range. The proximal transmission was therefore approximated by a gain-only model:
\begin{equation}
\label{eq:prox_pulley}
T_{\mathrm{out}} = \lambda^{\mathrm{prox}} T_{\mathrm{in}}.
\end{equation}
The distal coefficients were then derived by combining the identified proximal gain with the total gain and bias in \eqref{eq:pulley_tot}, using the algebraic cascade under Assumption~3:
\begin{equation}
\label{eq:pulley_cascade}
\begin{aligned}
T_{\mathrm{1}} &= \lambda^{\mathrm{prox}} T_{\mathrm{in}}, \\
T_{\mathrm{2}} &= \lambda^{\mathrm{dist}} T_{\mathrm{1}} + \delta^{\mathrm{dist}} = \lambda^{\mathrm{dist}}(\lambda^{\mathrm{prox}} T_{\mathrm{in}}) + \delta^{\mathrm{dist}}, \\
T_{\mathrm{out}} &= \lambda^{\mathrm{prox}} T_{\mathrm{2}} = \lambda^{\mathrm{dist}} (\lambda^{\mathrm{prox}})^2 T_{\mathrm{in}} + \lambda^{\mathrm{prox}} \delta^{\mathrm{dist}}.
\end{aligned}
\end{equation}
Matching the gain and bias in \eqref{eq:pulley_tot} and \eqref{eq:pulley_cascade} yields
\begin{equation} \label{eq:distal_params}
\lambda^{\mathrm{dist}} = \frac{\lambda^{\mathrm{tot}}}{(\lambda^{\mathrm{prox}})^2}, \quad \delta^{\mathrm{dist}} = \frac{\delta^{\mathrm{tot}}}{\lambda^{\mathrm{prox}}}.
\end{equation}

Fig.~\ref{fig:pulley_id} shows the measured tensions of both setups with the identified models. The resulting coefficients are $\lambda^{\mathrm{prox}}=0.4867$, $\lambda^{\mathrm{dist}}=3.2443$, and $\delta^{\mathrm{dist}}=1.7571$~N during pulling, and $\lambda^{\mathrm{prox}}=0.5400$, $\lambda^{\mathrm{dist}}=4.0418$, and $\delta^{\mathrm{dist}}=3.9020$~N during releasing. Note that $\lambda^{\mathrm{prox}}$ and $\lambda^{\mathrm{dist}}$ are effective pulley transmission coefficients obtained from the measured transmission relationships under Assumption~3. These coefficients describe the tension transmission effects of the pulley assembly and do not directly represent the physical efficiency of each pulley.  These calibrated pulley coefficients were used for parameter mapping across all configurations in both mapping validation and tension-tracking experiments, without configuration-specific adjustment.

\subsection{PRTL Tension Model for Actuation TSM Parameter Estimation}
\label{subsec:prtl_model}

Fig.~\ref{fig:prtl_model}(a) shows the shared routing of the PRTL sheath passes and the actuation TSM. Under Assumptions~1 and~2, each PRTL sheath pass shares the accumulated bending angle and mechanical parameters of the actuation TSM. Under these assumptions, each PRTL sheath pass has the same propagation gain and bias as the actuation TSM, denoted by $\Gamma^{\mathrm{act}}$ and $\beta^{\mathrm{act}}$. From the corrected capstan relation in \eqref{eq:capstan_pass},
\begin{equation}
\label{eq:capstan_params}
\Gamma^{\mathrm{act}} = e^{-(\mu\zeta + R')\Phi}, \qquad \beta^{\mathrm{act}} = A\,(\Gamma^{\mathrm{act}} - 1).
\end{equation}
The input-to-output coordinate follows the PRTL loop, so both passes have the same sliding sign $\zeta$.

Under Assumption~3, the five stages in Fig.~\ref{fig:prtl_model}(b) are composed using the pulley parameters identified in Section~\ref{sec:pre-def}. Let $T_{\mathrm{out}}^{i}$ ($i=1,\ldots,4$) denote the intermediate tensions. The resulting cascade is
\begin{equation}
\label{eq:stage_chain}
\begin{aligned}
T_{\mathrm{out}}^{1} &= \lambda^{\mathrm{prox}}T_{\mathrm{in}}^{\mathrm{prtl}}, & & \\
T_{\mathrm{out}}^{2} &= \Gamma^{\mathrm{act}} T_{\mathrm{out}}^{1} + \beta^{\mathrm{act}}, &
T_{\mathrm{out}}^{3} &= \lambda^{\mathrm{dist}} T_{\mathrm{out}}^{2} + \delta^{\mathrm{dist}}, \\
T_{\mathrm{out}}^{4} &= \Gamma^{\mathrm{act}} T_{\mathrm{out}}^{3} + \beta^{\mathrm{act}}, &
T_{\mathrm{out}}^{\mathrm{prtl}} &= \lambda^{\mathrm{prox}} T_{\mathrm{out}}^{4}.
\end{aligned}
\end{equation}
Successive substitution gives the modeled PRTL output tension:
\begin{equation}
\label{eq:t_out_5}
\begin{aligned}
T_{\mathrm{out}}^{\mathrm{prtl}}
&= (\Gamma^{\mathrm{act}})^2 (\lambda^{\mathrm{prox}})^2 \lambda^{\mathrm{dist}} T_{\mathrm{in}}^{\mathrm{prtl}} \\
&\quad + \lambda^{\mathrm{prox}}\bigl[(1+\Gamma^{\mathrm{act}}\lambda^{\mathrm{dist}})\beta^{\mathrm{act}}
+ \Gamma^{\mathrm{act}}\delta^{\mathrm{dist}}\bigr].
\end{aligned}
\end{equation}

Matching the gain and bias of this cascade to the measured relation $T_{\mathrm{out}}^{\mathrm{prtl}}=\Gamma^{\mathrm{prtl}}T_{\mathrm{in}}^{\mathrm{prtl}}+\beta^{\mathrm{prtl}}$ yields
\begin{equation}
\label{eq:prtl_mapping}
\begin{aligned}
\Gamma^{\mathrm{prtl}} &\approx (\Gamma^{\mathrm{act}})^2 (\lambda^{\mathrm{prox}})^2 \lambda^{\mathrm{dist}}, \\
\beta^{\mathrm{prtl}} &\approx \lambda^{\mathrm{prox}}\bigl[(1+\Gamma^{\mathrm{act}}\lambda^{\mathrm{dist}})\beta^{\mathrm{act}}
+ \Gamma^{\mathrm{act}}\delta^{\mathrm{dist}}\bigr].
\end{aligned}
\end{equation}

Inverting these relations yields the actuation TSM parameter estimates:
\begin{equation}
\label{eq:prtl_inverse}
\begin{aligned}
\hat{\Gamma}^{\mathrm{act}}
&= \sqrt{\frac{\Gamma^{\mathrm{prtl}}}{(\lambda^{\mathrm{prox}})^2\lambda^{\mathrm{dist}}}}, \\
\hat{\beta}^{\mathrm{act}}
&= \frac{\beta^{\mathrm{prtl}} - \lambda^{\mathrm{prox}}\hat{\Gamma}^{\mathrm{act}}\delta^{\mathrm{dist}}}
{\lambda^{\mathrm{prox}}\bigl(1+\hat{\Gamma}^{\mathrm{act}}\lambda^{\mathrm{dist}}\bigr)}.
\end{aligned}
\end{equation}

This mapping is based on an approximate model that uses effective pulley coefficients obtained from the calibration in Section~\ref{sec:pre-def}, assumes shared sheath transmission parameters, and neglects the proximal pulley bias. The positive root is selected for the propagation gain, and the mapping is applied separately using pulling or releasing parameters. The estimates are used in the feedforward controller in Fig.~\ref{fig:block_diagram}.


\section{Hysteresis Compensation Control}
\label{sec:control}
 
This section develops the PRTL-based hysteresis feedforward compensation controller using the mapped propagation parameters from~\Cref{sec:mapping}.

\subsection{Propagation Phase Compensation}
\label{subsec:prop_comp}
 
Based on the propagation relation in~\eqref{eq:Tension_base},
the propagation phases are parameterized 
by a slope $\Gamma_{\ell}^{\mathrm{act}}$ and a bias $\beta_{\ell}^{\mathrm{act}}$
for each direction.
Using the estimated slopes and biases, the forward model
for each propagation direction is written as
\begin{equation}
\label{eq:forward_model}
T_{\mathrm{out}} =
\begin{cases}
  \hat{\Gamma}_{\mathrm{p}}^{\mathrm{act}}\,T_{\mathrm{in}}
    + \hat{\beta}_{\mathrm{p}}^{\mathrm{act}}, & \text{(PP)} \\[4pt]
  \hat{\Gamma}_{\mathrm{r}}^{\mathrm{act}}\,T_{\mathrm{in}}
    + \hat{\beta}_{\mathrm{r}}^{\mathrm{act}}, & \text{(RP)}
\end{cases}
\end{equation}
Here, $\hat{\Gamma}_{\mathrm{p}}^{\mathrm{act}}$, $\hat{\Gamma}_{\mathrm{r}}^{\mathrm{act}}$ and $\hat{\beta}_{\mathrm{p}}^{\mathrm{act}}$, $\hat{\beta}_{\mathrm{r}}^{\mathrm{act}}$ are the slopes and biases, respectively, estimated from the PRTL loop parameters through \eqref{eq:prtl_inverse}.

To implement the model-based feedforward compensation, the feedforward command is derived by inverting the forward model \eqref{eq:forward_model} : 
\begin{equation}
\label{eq:ff_cmd}
T_{\mathrm{cmd},\,\ell}
  = \frac{T_{\mathrm{ref}} - \hat{\beta}_{\ell}^{\mathrm{act}}}
         {\hat{\Gamma}_{\ell}^{\mathrm{act}}},
  \quad \ell \in \{\mathrm{p},\,\mathrm{r}\}.
\end{equation}
 where $T_{\mathrm{cmd},\,\ell}$ is the required input tension command for the output tension $T_{\mathrm{out}}$ to track the reference tension $T_{\mathrm{ref}}$, and the subscript $\ell$ denotes the direction of the tendon motion.

 \begin{figure}[t!]
    \centering
    \includegraphics[width=\columnwidth]{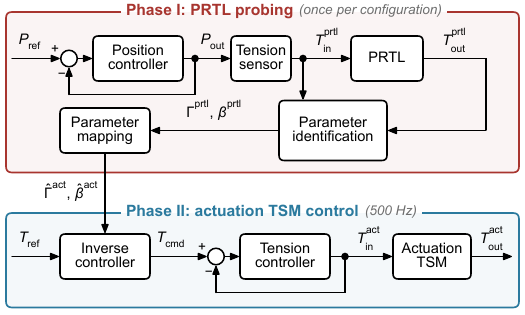}
    \caption{PRTL-based compensation control. Phase~I (red): proximal tension measurements during probing identify the PRTL parameters, which are mapped to actuation TSM parameters using \eqref{eq:prtl_inverse}. Phase~II (blue): the inverse controller generates $T_{\mathrm{cmd}}$, and the tension controller regulates the actuation input tension.}
    \vspace{-15pt}
    \label{fig:block_diagram}
\end{figure}

\subsection{Backlash Phase Compensation}
\label{subsec:bl_comp}
 
In the backlash phases, 
$T_{\mathrm{out}}$ does not vary with $T_{\mathrm{in}}$, 
and a functional input--output relationship 
cannot be established.
To compensate for this dead zone, 
the controller first identifies the backlash 
boundaries and then applies an error-based 
command to recover tension transmission.

\noindent 1) {\fontfamily{lmr}\selectfont\itshape Backlash Boundary Identification.}
 
Following the criterion in~\cite{TSMCOMP_FBG}, 
the backlash phase begins when 
the tension direction reverses.
At the reversal point, the measured input tension 
is recorded as $T_{\mathrm{in}}^{\mathrm{rev}}$, 
and the corresponding output tension is estimated as
\begin{equation}
\label{eq:Tout_rev}
T_{\mathrm{out}}^{\mathrm{rev}}
  = \hat{\Gamma}_{d}^{\mathrm{act}}\,
    T_{\mathrm{in}}^{\mathrm{rev}} + \hat{\beta}_{d}^{\mathrm{act}},
\end{equation}
where $d$ denotes the preceding propagation direction 
($d\!=\!\mathrm{r}$ after RP, $d\!=\!\mathrm{p}$ after PP).
This value remains constant throughout the backlash phase.
The backlash phase ends when the next propagation model 
intersects $T_{\mathrm{out}}^{\mathrm{rev}}$.
Setting 
$\hat{\Gamma}_{\ell}^{\mathrm{act}}\,T_{\mathrm{in}}(t)
 + \hat{\beta}_{\ell}^{\mathrm{act}} = T_{\mathrm{out}}^{\mathrm{rev}}$
and solving for $T_{\mathrm{in}}$ gives 
the transition thresholds:
\begin{equation}
\label{eq:bl_threshold}
\begin{aligned}
T_{\mathrm{th},\mathrm{p}}
  &= \frac{\hat{\Gamma}_{\mathrm{r}}^{\mathrm{act}}\,
           T_{\mathrm{in}}^{\mathrm{rev}}
           + (\hat{\beta}_{\mathrm{r}}^{\mathrm{act}} - \hat{\beta}_{\mathrm{p}}^{\mathrm{act}})}
          {\hat{\Gamma}_{\mathrm{p}}^{\mathrm{act}}}, \\[4pt]
T_{\mathrm{th},\mathrm{r}}
  &= \frac{\hat{\Gamma}_{\mathrm{p}}^{\mathrm{act}}\,
           T_{\mathrm{in}}^{\mathrm{rev}}
           + (\hat{\beta}_{\mathrm{p}}^{\mathrm{act}} - \hat{\beta}_{\mathrm{r}}^{\mathrm{act}})}
          {\hat{\Gamma}_{\mathrm{r}}^{\mathrm{act}}}.
\end{aligned}
\end{equation}
The PB phase ends when 
$T_{\mathrm{in}}(t) \ge T_{\mathrm{th},\mathrm{p}}$.
The RB phase ends when 
$T_{\mathrm{in}}(t) \le T_{\mathrm{th},\mathrm{r}}$.

\noindent 2) {\fontfamily{lmr}\selectfont\itshape Backlash Command.}
 
The tracking error during backlash is 
$\Delta T = T_{\mathrm{ref}} - T_{\mathrm{out}}^{\mathrm{rev}}$.
This error is added to the reference 
to drive the tendon through the dead zone:
\begin{equation}
\label{eq:bl_cmd1}
\tilde{T}_{\mathrm{ref}}
  = T_{\mathrm{ref}} + \Delta T
  = 2\,T_{\mathrm{ref}} - T_{\mathrm{out}}^{\mathrm{rev}}.
\end{equation}
The intermediate reference $\tilde{T}_{\mathrm{ref}}$ 
is then passed through the inverse propagation model:
\begin{equation}
\label{eq:bl_cmd2}
T_{\mathrm{cmd},\,\ell}
  = \frac{\tilde{T}_{\mathrm{ref}}
          - \hat{\beta}_{\ell}^{\mathrm{act}}}
         {\hat{\Gamma}_{\ell}^{\mathrm{act}}},
  \quad \ell \in \{\mathrm{p},\,\mathrm{r}\}.
\end{equation}

\begin{figure}[t!]
    \centering
    \includegraphics[width=0.85\columnwidth]{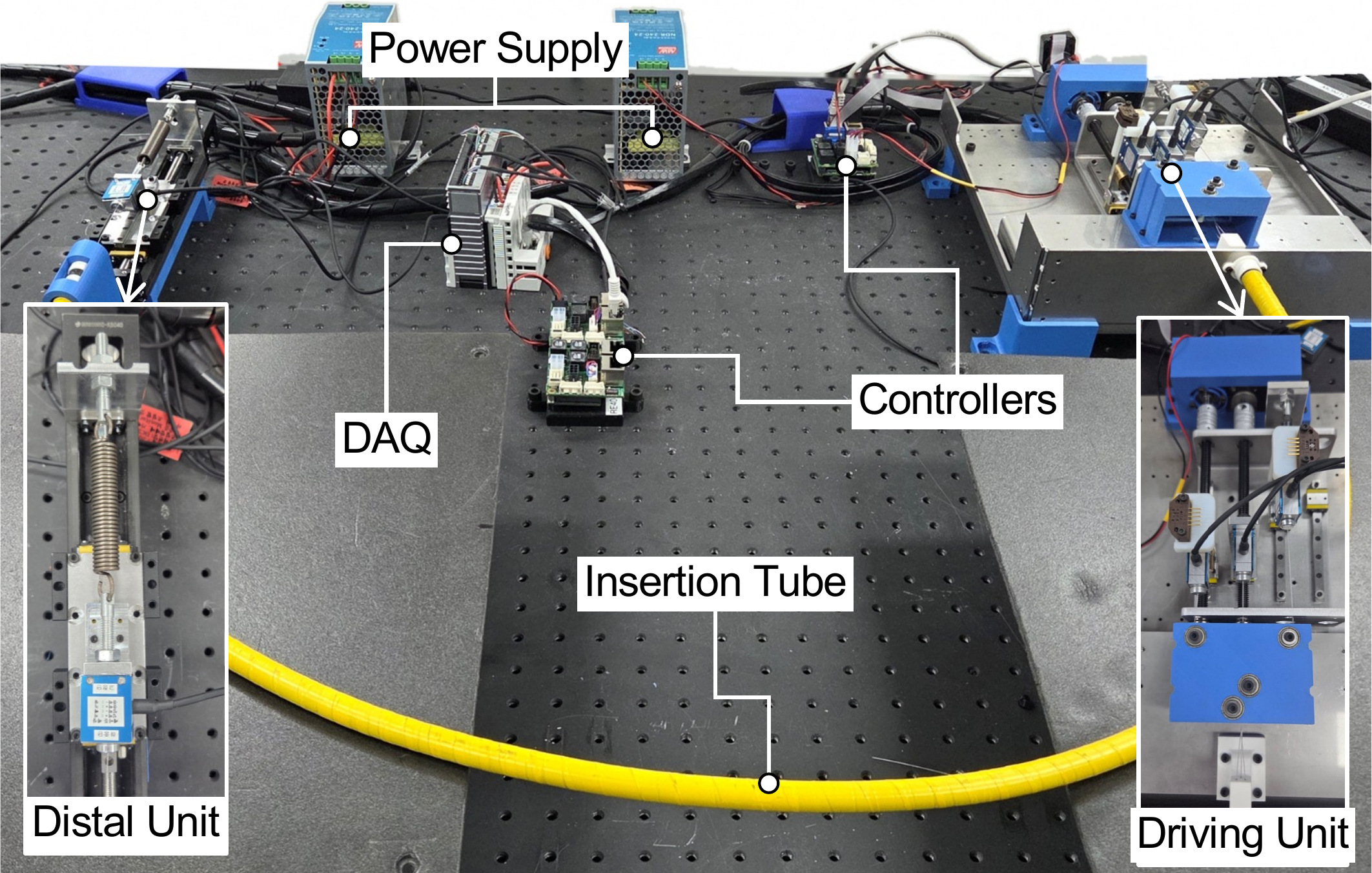}
    \caption{Experimental platform with enlarged views of the distal module and proximal driving unit.}
    \vspace{-15pt}
    \label{fig:system_configuration}
\end{figure}

  \begin{figure*}[t]
    \centering
    \includegraphics[width=\textwidth]{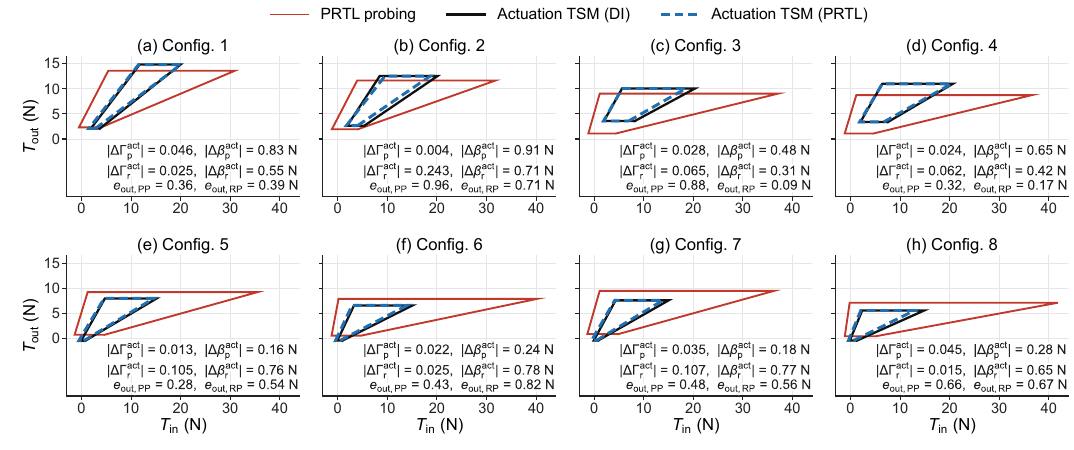}
    \caption{Parameter mapping validation across eight bending configurations. Red: PRTL probing; black: actuation TSM identified directly (DI); dashed blue: actuation TSM predicted from PRTL parameters. The two actuation loops use the same output-tension limits. Annotations show absolute gain and bias errors relative to DI and output-tension RMSEs for pulling and releasing propagation.}
    \label{fig:model_validation}
\end{figure*}

\begin{table*}[t]
\centering
\caption{Errors of PRTL-Based Parameter Mapping over the Eight Bending Configurations}
\label{tab:mapping_validation}
\footnotesize
\setlength{\tabcolsep}{3.5pt}
\begin{tabular*}{\textwidth}{@{\extracolsep{\fill}}l cccc cc cc}
\toprule
 & \multicolumn{4}{c}{\makecell{Parameter error\\{\scriptsize RMSE (SD)}}} & \multicolumn{2}{c}{\makecell{Propagation output RMSE,\\{\scriptsize mean (SD)}}} & \multicolumn{2}{c}{\makecell{Backlash boundary RMSE,\\{\scriptsize mean (SD)}}} \\
\cmidrule(lr){2-5} \cmidrule(lr){6-7} \cmidrule(lr){8-9}
 & $\Gamma_{\mathrm{p}}^{\mathrm{act}}$ & $\beta_{\mathrm{p}}^{\mathrm{act}}$ (N) & $\Gamma_{\mathrm{r}}^{\mathrm{act}}$ & $\beta_{\mathrm{r}}^{\mathrm{act}}$ (N) & $e_{\mathrm{out},\mathrm{PP}}$ (N) & $e_{\mathrm{out},\mathrm{RP}}$ (N) & $e_{\mathrm{bnd},\mathrm{PB}}$ (N) & $e_{\mathrm{bnd},\mathrm{RB}}$ (N) \\
\midrule
Summary & 0.030 (0.015) & 0.54 (0.30) & 0.107 (0.074) & 0.64 (0.18) & 0.55 (0.26) & 0.49 (0.26) & 0.62 (0.22) & 0.81 (0.50) \\
Max. & 0.046 & 0.91 & 0.243 & 0.78 & 0.96 & 0.82 & 0.99 & 1.38 \\
Min. & 0.004 & 0.16 & 0.015 & 0.31 & 0.28 & 0.09 & 0.39 & 0.17 \\
\bottomrule
\end{tabular*}
\par\smallskip
\parbox{0.98\textwidth}{\footnotesize Parameter columns report RMSE (SD of absolute DI--PRTL differences); phase columns report mean (SD) of configuration-specific RMSEs. Max. and Min. give the extrema of absolute parameter errors or phase RMSEs across eight configurations.}
\end{table*}

\subsection{Integrated Compensation Procedure}
\label{subsec:integrated}
The integrated procedure consists of PRTL-based parameter identification and mapping, followed by feedforward compensation of the actuation TSM (Fig.~\ref{fig:block_diagram}). First, the PRTL executes the 10-s probing trajectory in Fig.~\ref{fig:sstl_design}(c) under PID position control. The measured input--output tensions are used to identify the PRTL parameters (Section~\ref{subsec:Parameter_Identification}), which are mapped to the actuation TSM parameters using \eqref{eq:prtl_inverse}.

During actuation, the inverse feedforward controller uses the mapped parameters and the backlash boundaries in \eqref{eq:bl_threshold} to select the compensation command: \eqref{eq:bl_cmd2} during backlash and \eqref{eq:ff_cmd} during propagation. A PID tension controller regulates the measured input tension to track $T_{\mathrm{cmd}}$, while the feedforward compensation accounts for hysteresis so that the output tension follows $T_{\mathrm{ref}}$.

\section{Experiments and Results}
This section validates both the PRTL parameter mapping and the resulting tension-tracking performance. Across eight bending configurations, the actuation TSM parameters estimated from PRTL measurements are compared with reference parameters obtained by direct identification (DI). The control experiment tests whether these estimates improve tracking relative to uncompensated control and approach DI-based compensation across four bending configurations under triangular and random references.

\subsection{Experimental Platform}
\label{sec:exp_setup}
Fig.~\ref{fig:system_configuration} shows the experimental platform. A Xenomai-based real-time PC controls the actuation TSM motor (XM430-W350, ROBOTIS) via RS-485 and the PRTL motor (RE25, Maxon) through an EPOS4 controller via EtherCAT, both at 1~kHz. Tensions are measured using load cells (333FDX, Ktoyo) and acquired at 500~Hz through a Beckhoff ELM3504-0000 terminal. The tension data used for evaluation are logged at 50~Hz. The 3D-printed distal pulley module is integrated with the insertion tube and mounted on the experimental support structure.

\subsection{Comparison Methods}
\label{sec:comp_methods}
The experiments evaluate parameter mapping accuracy and tension-tracking performance. Mapping accuracy is assessed against directly identified parameters. Tracking performance is compared among uncompensated control and feedforward compensation using either PRTL-mapped or directly identified parameters.

\noindent 1) {\fontfamily{lmr}\selectfont\itshape No compensation.} The actuation TSM is driven directly by the reference tension $T_{\mathrm{ref}}$, providing a baseline for the tracking error without hysteresis compensation.

\noindent 2) {\fontfamily{lmr}\selectfont\itshape Proposed (PRTL-based).} PRTL parameters are identified from the 10-s probing trajectory (Fig.~\ref{fig:sstl_design}(c)) and mapped to actuation TSM parameters using \eqref{eq:prtl_inverse}. These parameters are then used for feedforward tension compensation (Fig.~\ref{fig:block_diagram}).

\noindent 3) {\fontfamily{lmr}\selectfont\itshape Direct identification (DI).} Before the experiments in each configuration, a triangular tension reference is applied to the actuation TSM, and the parameters are identified directly from the measured proximal input tension $T_{\mathrm{in}}$ and distal output tension $T_{\mathrm{out}}$ using the procedure in \Cref{subsec:Parameter_Identification}. The reference ranges from 0 to 15~N with a period of 133~s for the tracking experiments and stays within 0--20~N for the mapping validation. These parameters provide the reference for mapping validation and are used by the same feedforward controller for the tracking comparison.

The proposed method does not require distal tension measurements for parameter estimation or control. In these experiments, distal measurements were used only to evaluate mapping and tension-tracking performance and to identify parameters for DI-based control.

\begin{figure*}[t]
    \centering
    \includegraphics[width=\textwidth]{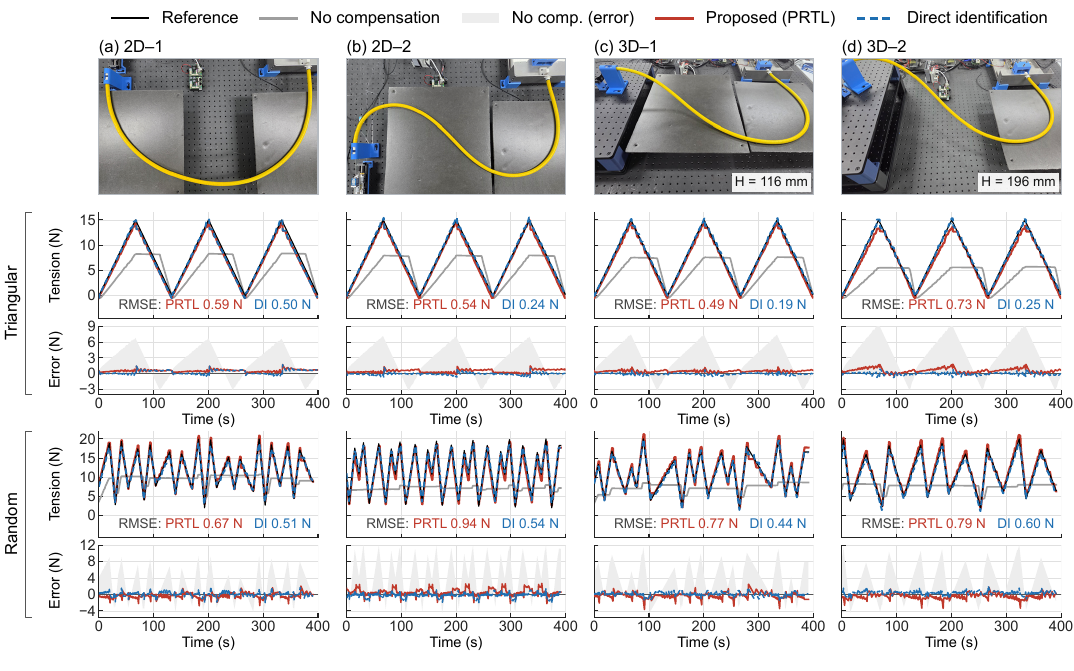}
    \caption{Tension tracking under triangular and random references. Columns (a)--(d) show 2D--1, 2D--2, 3D--1 ($H=116$~mm), and 3D--2 ($H=196$~mm). Each column contains the configuration photograph, tension responses, and tracking errors (reference minus output). Annotated PRTL and DI RMSEs use the full evaluation intervals, as in Table~\ref{tab:control_configurations}.}
    \label{fig:control_validation}
\end{figure*}

\subsection{Hysteresis Parameter Mapping Validation}
\label{subsec:estimation_validation}
The PRTL-mapped and DI-based actuation TSM models were compared across eight configurations (Fig.~\ref{fig:model_validation}). Output tensions were predicted using \eqref{eq:forward_model}. The evaluation examined parameter errors, propagation output-tension errors, and backlash boundary errors. PRTL parameters were identified from the 10-s probing trajectory (Fig.~\ref{fig:sstl_design}(c)).

\noindent 1) {\fontfamily{lmr}\selectfont\itshape Parameter Errors.}
Relative to DI, the RMSEs for $\Gamma_{\mathrm{p}}^{\mathrm{act}}$ and $\Gamma_{\mathrm{r}}^{\mathrm{act}}$ were 0.030 and 0.107, respectively, and those for $\beta_{\mathrm{p}}^{\mathrm{act}}$ and $\beta_{\mathrm{r}}^{\mathrm{act}}$ were 0.54~N and 0.64~N (\Cref{tab:mapping_validation}). The RMSE was larger for $\Gamma_{\mathrm{r}}^{\mathrm{act}}$ than for $\Gamma_{\mathrm{p}}^{\mathrm{act}}$. On average, $\beta_{\mathrm{p}}^{\mathrm{act}}$ and $\beta_{\mathrm{r}}^{\mathrm{act}}$ were overestimated by 0.47~N and 0.62~N, respectively. Fig.~\ref{fig:model_validation} reports the absolute parameter errors for each configuration.

\noindent 2) {\fontfamily{lmr}\selectfont\itshape Propagation Output-Tension Error.}
Propagation accuracy is evaluated by comparing the output tensions predicted by the PRTL-mapped and DI-based actuation TSM models at the same input tension. Let $f_{j,k}^{s}(u)$ denote the forward propagation relation in \eqref{eq:forward_model} for input tension $u$, phase $j\in\{\mathrm{PP},\mathrm{RP}\}$, and configuration $k$, using parameters from $s\in\{\mathrm{DI},\mathrm{PRTL}\}$. The evaluation interval $[a_{j,k},b_{j,k}]$ is defined by the DI branch at the minimum and maximum measured actuation output tensions of the displayed cycle. The output-tension RMSE is
\begin{equation}
\label{eq:propagation_output_err}
e_{\mathrm{out},j,k}=\sqrt{\frac{1}{b_{j,k}-a_{j,k}}\int_{a_{j,k}}^{b_{j,k}}\left[f_{j,k}^{\mathrm{PRTL}}(u)-f_{j,k}^{\mathrm{DI}}(u)\right]^2\,du}.
\end{equation}
Across the eight configurations, $e_{\mathrm{out},\mathrm{PP},k}$ and $e_{\mathrm{out},\mathrm{RP},k}$ were $0.55\pm0.26$~N and $0.49\pm0.26$~N, respectively (\Cref{tab:mapping_validation}). Fig.~\ref{fig:model_validation} reports both errors for each configuration.

\noindent 3) {\fontfamily{lmr}\selectfont\itshape Backlash Boundary Error.}
Backlash accuracy is evaluated from the positions of the two input-tension boundaries of each backlash interval. PB and RB are evaluated at the same lower and upper output-tension limits, respectively, in both models. The intersections of the pulling and releasing branches with each output limit define the interval endpoints $a_{j,k}^{s}\leq b_{j,k}^{s}$, where $j\in\{\mathrm{PB},\mathrm{RB}\}$. Their RMSE relative to DI is
\begin{equation}
\label{eq:backlash_boundary_err}
e_{\mathrm{bnd},j,k}=\sqrt{\frac{\left(a_{j,k}^{\mathrm{PRTL}}-a_{j,k}^{\mathrm{DI}}\right)^2+\left(b_{j,k}^{\mathrm{PRTL}}-b_{j,k}^{\mathrm{DI}}\right)^2}{2}}.
\end{equation}
This metric captures both changes in backlash length and horizontal translation in input tension. Across the eight configurations, $e_{\mathrm{bnd},\mathrm{PB},k}$ and $e_{\mathrm{bnd},\mathrm{RB},k}$ were $0.62\pm0.22$~N and $0.81\pm0.50$~N, respectively (\Cref{tab:mapping_validation}).

Across the eight tested configurations, the mean discrepancies relative to DI were approximately 0.5~N in propagation output tension and 0.6--0.8~N in backlash boundary location along the input-tension axis. To evaluate how effectively the mapped parameters support tension control, the following section presents tension-tracking experiments using these parameters for feedforward compensation.

\begin{table*}[t]
\centering
\caption{Tension Tracking Performance by Configuration and Reference Trajectory}
\label{tab:control_configurations}
\fontfamily{lmr}\selectfont
\renewcommand{\bfdefault}{bx}
\renewcommand{\arraystretch}{1.12}
\setlength{\tabcolsep}{5pt}
\begin{tabular*}{\textwidth}{@{\extracolsep{\fill}}llcccccc}
\toprule
 & & \multicolumn{3}{c}{RMSE (error SD), N} & \multicolumn{3}{c}{WAPE (\%)} \\
\cmidrule(lr){3-5}\cmidrule(lr){6-8}
Trajectory & Configuration & No comp. & DI & \textbf{Proposed} & No comp. & DI & \textbf{Proposed} \\
\midrule
\multirow{5}{*}{Triangular}
 & 2D--1 & 3.60 (2.80) & 0.50 (0.39) & \textbf{0.59 (0.23)} & 40.81 & 5.06 & \textbf{7.27} \\
 & 2D--2 & 3.80 (2.91) & 0.24 (0.24) & \textbf{0.54 (0.30)} & 43.14 & 2.27 & \textbf{6.15} \\
 & 3D--1 & 3.97 (3.00) & 0.19 (0.18) & \textbf{0.49 (0.19)} & 44.99 & 1.95 & \textbf{5.98} \\
 & 3D--2 & 5.12 (3.47) & 0.25 (0.25) & \textbf{0.73 (0.43)} & 57.94 & 2.55 & \textbf{8.22} \\
\cmidrule(lr){2-8}
 & Average & 4.12 (0.68) & 0.30 (0.14) & \textbf{0.59 (0.11)} & 46.72 (7.67) & 2.96 (1.42) & \textbf{6.91 (1.05)} \\
\midrule
\multirow{5}{*}{Random}
 & 2D--1 & 4.18 (3.42) & 0.51 (0.51) & \textbf{0.67 (0.59)} & 29.06 & 3.29 & \textbf{4.52} \\
 & 2D--2 & 6.31 (4.34) & 0.54 (0.52) & \textbf{0.94 (0.74)} & 45.91 & 3.21 & \textbf{6.24} \\
 & 3D--1 & 5.08 (3.53) & 0.44 (0.43) & \textbf{0.77 (0.72)} & 38.44 & 2.96 & \textbf{5.39} \\
 & 3D--2 & 5.20 (3.92) & 0.60 (0.42) & \textbf{0.79 (0.65)} & 38.17 & 4.70 & \textbf{5.68} \\
\cmidrule(lr){2-8}
 & Average & 5.19 (0.87) & 0.52 (0.07) & \textbf{0.79 (0.11)} & 37.90 (6.89) & 3.54 (0.79) & \textbf{5.46 (0.72)} \\
\bottomrule
\end{tabular*}
\par\smallskip
\begin{minipage}{0.98\textwidth}
\footnotesize For each configuration, the value in parentheses is the sample SD of the tracking error within the trajectory. In the Average rows, it is the sample SD across the four configurations. Boldface highlights the proposed method.
\end{minipage}
\end{table*}

\subsection{Tension Tracking Performance}
\label{sec:results}
The three control methods were evaluated under triangular and random tension references in four bending configurations: two planar (2D--1 and 2D--2) and two spatial (3D--1 and 3D--2). Fig.~\ref{fig:control_validation} shows the measured responses, and \Cref{tab:control_configurations} summarizes the performance.

For each configuration, RMSE and the standard deviation (SD) of the pointwise tracking error, $T_{\mathrm{ref}}[k]-T_{\mathrm{out}}[k]$, are computed over the reference trajectory. SD measures the variation about the mean tracking error within that trajectory. The weighted absolute percentage error (WAPE) is defined as
\begin{equation}
\label{eq:wape}
  \mathrm{WAPE}
  = 100\frac{\sum_{k=1}^{N}\lvert T_{\mathrm{ref}}[k]
      - T_{\mathrm{out}}[k] \rvert}
         {\sum_{k=1}^{N}\lvert T_{\mathrm{ref}}[k]\rvert}
  \;\;(\%).
\end{equation}
Here, $N$ is the total number of evaluation samples. WAPE expresses the accumulated absolute deviation of the output tension from the reference as a percentage of the accumulated reference magnitude; a lower value indicates closer tracking.

\noindent 1) {\fontfamily{lmr}\selectfont\itshape Triangular Reference.}
The triangular reference ranges from 0 to 15~N with a period of approximately 133~s. The metrics were computed over three cycles (399~s, 19,950 samples).
PRTL-based compensation reduced mean RMSE from 4.12~N under uncompensated control to 0.59~N, an 85.7\% reduction. Mean WAPE decreased from 46.72\% to 6.91\%. DI-based compensation achieved a mean RMSE of 0.30~N and WAPE of 2.96\%.

\noindent 2) {\fontfamily{lmr}\selectfont\itshape Random Reference.}
The random reference is generated by combining sinusoidal components with frequencies ranging from 0.02 to 0.06~Hz and ranges from 2 to 20~N. A different random trajectory was used for each configuration, with the same trajectory applied to all three methods within a configuration. The metrics were computed over 391~s (19,550 samples). PRTL-based compensation reduced mean RMSE from 5.19~N to 0.79~N, an 84.8\% reduction. Mean WAPE decreased from 37.90\% to 5.46\%. DI-based compensation achieved a mean RMSE of 0.52~N and WAPE of 3.54\%.

The mean tracking RMSEs of PRTL-based compensation were close to those of DI-based compensation, with differences of 0.29~N and 0.27~N under triangular and random references, respectively, while those of uncompensated control differed from DI-based compensation by 3.82~N and 4.67~N. This comparison supports the use of PRTL-mapped parameters for effective tension compensation without distal sensing for parameter estimation.

\section{Conclusion}
\label{sec:conclusion}
This study presented the PRTL for estimating configuration-dependent actuation TSM transmission parameters. A 10-s probing motion characterized the PRTL tension transmission without driving the actuation tendon. A closed-form mapping accounted for the two sheath passes and calibrated pulley transmission to estimate actuation TSM parameters. These parameters were used for feedforward compensation during propagation and backlash.

Across eight bending configurations, the PRTL-mapped actuation TSM models yielded mean RMSEs of 0.49--0.55~N for output-tension predictions and 0.62--0.81~N for backlash boundary locations relative to the DI-based models. 
PRTL-based compensation reduced mean tracking RMSE by 85.7\% and 84.8\% relative to uncompensated control, with gaps of 0.29~N and 0.27~N from DI-based compensation for triangular and random references, respectively. These results support proximal PRTL probing and parameter mapping for effective tension compensation.

The mapping assumes similar bending geometry and mechanical properties among the co-routed TSMs, and validation was limited to a single actuation TSM. The experiments evaluated tension transmission identification and compensation; distal joint motion and tip-position accuracy were not assessed. Future work will focus on extending the method to multiple actuation TSMs and integrating it into endoscopic surgical robots.


\bibliographystyle{Bibliography/IEEEtranTIE}
\bibliography{Bibliography/IEEEabrv,ref}

\begin{IEEEbiography}[{\includegraphics[width=1in,height=1.25in,clip,keepaspectratio]{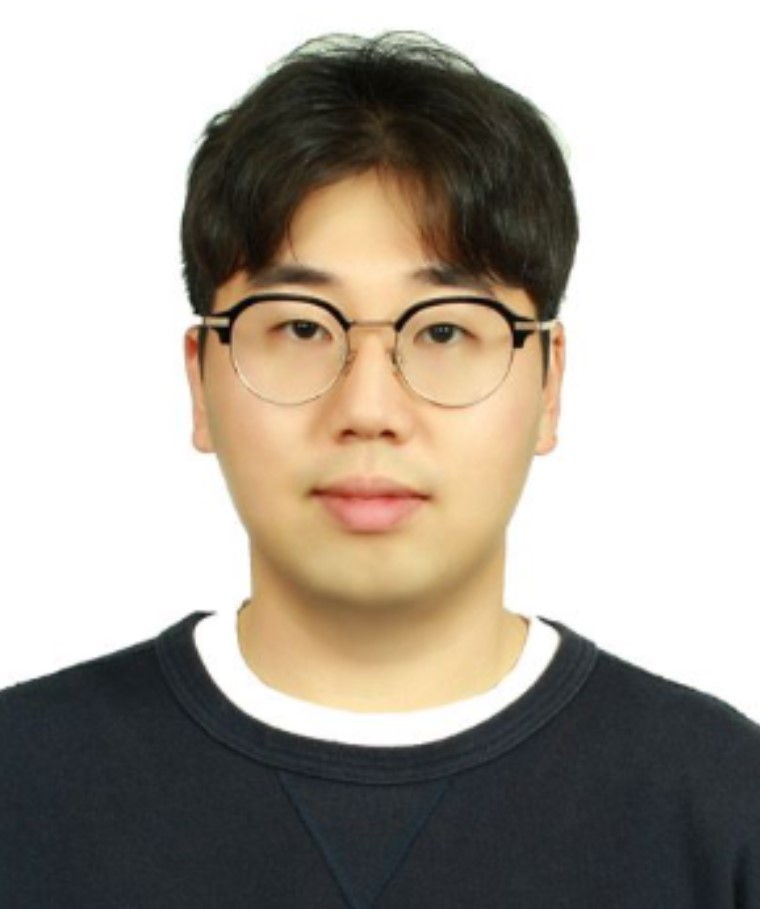}}]
{Myeongbo Park} received the B.S. degree in electronic engineering from Kyungpook National University, Daegu, Republic of Korea, in 2023. He is currently working toward the Ph.D. degree in robotics and mechatronics engineering with the Daegu Gyeongbuk Institute of Science and Technology (DGIST), Daegu, Republic of Korea.
His research interests include compensation of tendon--sheath mechanisms for surgical continuum robots and continuum robot dynamics modeling.
\end{IEEEbiography}

\begin{IEEEbiography}[{\includegraphics[width=1in,height=1.25in,clip,keepaspectratio]{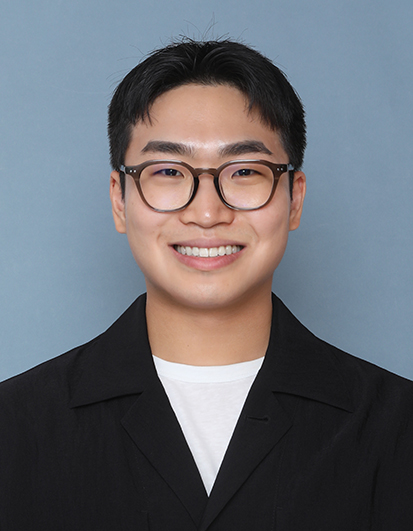}}]
{Junhyun Park} received the B.E. degree in computer science and electrical engineering and the M.S. degree in robotics and mechatronics engineering from the Daegu Gyeongbuk Institute of Science and Technology (DGIST), Daegu, Republic of Korea, in 2024 and 2026, respectively. He is currently a Researcher with the AI Research Laboratory, DEEPNOID, Seoul, Republic of Korea. His research interests include continuum manipulators, tendon-driven mechanisms, hysteresis compensation, surgical automation, and AI-driven medical image analysis. He was a recipient of the DGIST Presidential Fellowship (2021--2024), the Korea Presidential Science Scholarship (2022--2024), and the Talent Award of Korea from the Ministry of Education (2025).
\end{IEEEbiography}

\begin{IEEEbiography}[{\includegraphics[width=1in,height=1.25in,clip,keepaspectratio]{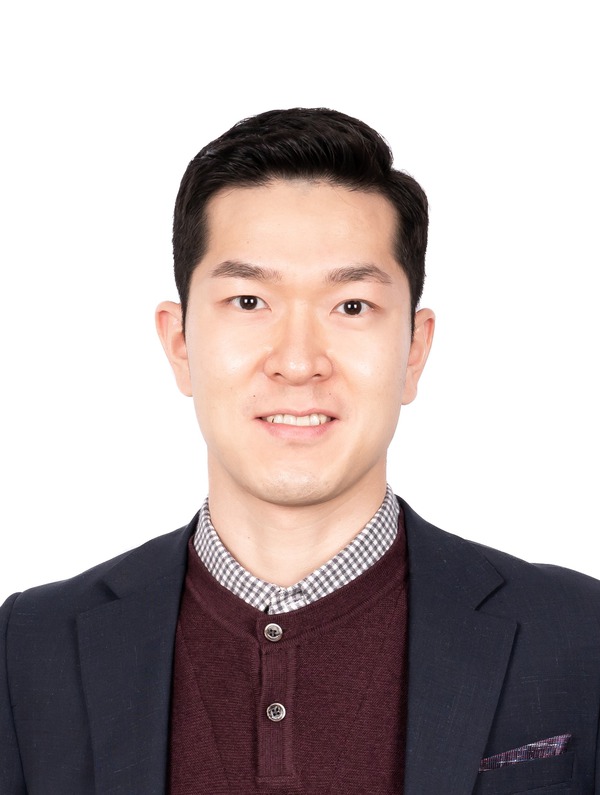}}]
{Minho Hwang} received the B.S., M.S., and Ph.D. degrees in mechanical engineering from the Korea Advanced Institute of Science and Technology (KAIST), Daejeon, Republic of Korea, in 2010, 2012, and 2017, respectively. From 2019 to 2021, he was a Postdoctoral Research Associate with the Department of Electrical Engineering and Computer Sciences, University of California, Berkeley, CA, USA. He is currently an Associate Professor with the Department of Robotics and Mechatronics Engineering, Daegu Gyeongbuk Institute of Science and Technology (DGIST), Daegu, Republic of Korea. His research interests include surgical robotic systems, tendon-driven actuation and sensing, flexible robotic mechanisms, precision manipulation, and intelligent robot control.
\end{IEEEbiography}

\end{document}